\documentclass{article} 
\usepackage{iclr2021_conference,times}

\usepackage{amsmath,amsfonts,bm}

\def\eqref#1{equation~\ref{#1}}

\def\1{\bm{1}}

\DeclareMathAlphabet{\mathsfit}{\encodingdefault}{\sfdefault}{m}{sl}
\SetMathAlphabet{\mathsfit}{bold}{\encodingdefault}{\sfdefault}{bx}{n}

\usepackage[font=small,skip=0pt]{caption}

\usepackage{xcolor}
\PassOptionsToPackage{hyphens}{url}\usepackage[pagebackref=true,breaklinks=true,letterpaper=true,colorlinks,
  citecolor=citecolor,bookmarks=false]{hyperref}
\definecolor{citecolor}{RGB}{34,139,34}

\usepackage{url}
\usepackage{subfig}
\usepackage{graphicx}
\usepackage{multirow}
\usepackage{booktabs}
\usepackage{sidecap}

\title{Cooperating RPN's Improve Few-Shot Object Detection}

\author{Weilin Zhang, Yu-Xiong Wang,  David A. Forsyth \\
Department of Computer Science\\
University of Illinois at Urbana-Champaign\\
\texttt{\{weilinz2, yxw, daf\}@illinois.edu} \\
}

\newcommand{\RNum}[1]{\uppercase\expandafter{\romannumeral #1\relax}}

\iclrfinalcopy 
\begin{document}

\maketitle

\begin{abstract}
Learning to detect an object in an image from very few training examples - few-shot object detection - is challenging, because the classifier that sees proposal boxes has very little training data.  A particularly challenging training regime occurs when there are one or two training examples. In this case, if the region proposal network (RPN) misses even one high intersection-over-union (IOU) training box, the classifier's model of how object appearance varies can be severely impacted. We use multiple distinct yet cooperating RPN's.  Our RPN's are trained to be different, but not too different; doing so yields significant performance improvements over state of the art for COCO and PASCAL VOC in the very few-shot setting. This effect appears to be independent of the choice of classifier or dataset.  
\end{abstract}

\section{Introduction}
\label{sec:intro}
Achieving accurate few-shot object detection is difficult, because one must rely on a classifier building a useful model of variation in appearance with very few examples.  This paper identifies an important effect that causes existing detectors to have weaker than necessary performance in the few-shot regime. By remediating this difficulty, we obtain substantial improvements in performance with current architectures.

The effect is most easily explained by looking at the ``script'' that modern object detectors mostly follow. As one would expect, there are variations in detector structure, but these do not mitigate the effect. A modern object detector will first find promising image locations; these are usually, but not always, boxes. We describe the effect in terms of boxes reported by a region proposal network (RPN) \citep{ren2015faster}, but expect that it applies to other representations of location, too. The detector then passes the promising locations through a classifier to determine what, if any, object is present. Finally, it performs various cleanup operations (non-maximum suppression, bounding box regression, etc.), aimed at avoiding multiple predictions in the same location and improving localization. The evaluation procedure for reported labeled boxes uses an intersection-over-union (IOU) test as part of determining whether a box is relevant.

\begin{figure}[h]
    \centering
    \subfloat[TFA]{{\includegraphics[width=6cm]{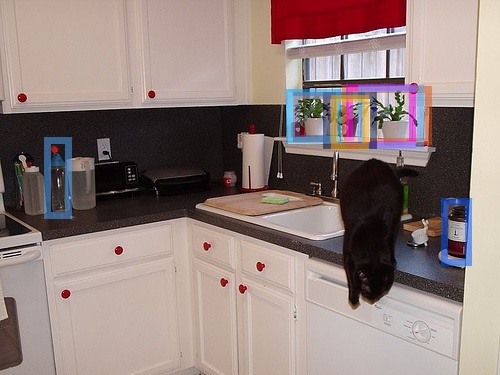} }}
    \qquad
    \subfloat[Ours]{{\includegraphics[width=6cm]{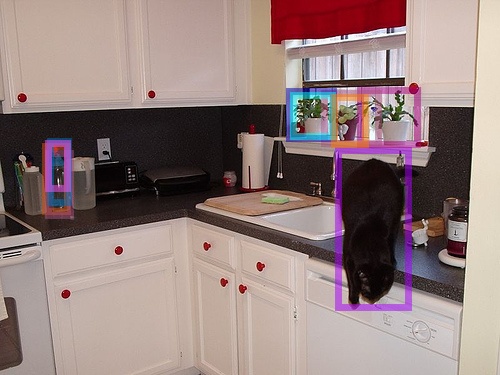} }}
    \caption{Few-shot detectors are subject to the {\em proposal neglect} effect.  On the {\bf left}: an image (from the PASCAL VOC split 3 novel classes test set) showing the top 10 proposals from a state-of-the-art few-shot detector TFA \citep{wang2020few}. On the {\bf right}: the top 10 proposals from our cooperating RPN's. Cat is in the novel classes \{boat, \textbf{cat}, mbike, sheep, sofa\} (i.e., both models are \textbf{cat} detectors). Note that the classifier in TFA {\em will not see a cat box}, and so it cannot detect the cat.  This is a disaster when this image is used in training.  For our approach, the cat box is found and is high in the top 10 proposals.  
    \label{fig:example}}
\end{figure}

A detector that is trained for few-shot detection is trained on two types of categories. Base categories have many training examples, and are used to train an RPN and the classifier. Novel categories have one (or three, or five, etc.) examples per category. The existing RPN is used to find boxes, and the classifier is fine-tuned to handle novel categories. 

Now assume that the detector must learn to detect a category from a single example. The RPN is already trained on other examples. It produces a collection of relevant boxes, which are used to train the classifier. The only way that the classifier can build a model of the categories variation in appearance is by having multiple high IOU boxes reported
by the RPN. In turn, this means that an RPN that behaves well on base categories may create serious problems for novel categories. Imagine that the RPN reports only a few of the available high IOU boxes in training data. For base categories, this is not a problem; many high IOU  boxes will pass to the classifier because there is a lot of training data, and so the classifier will be able to build a model of the categories variation in appearance. This variation will be caused by effects like aspect, in-class variation, and the particular RPN's choice of boxes.  But for novel categories, an RPN must report as many high IOU boxes as possible, because otherwise the classifier will have too weak a model of appearance variation -- for example, it might think that the object must be centered in the box. This will 
significantly depress accuracy. As Figure~\ref{fig:example} and our results illustrate, this effect (which we call {\bf proposal neglect}) is present in the state-of-the-art few-shot detectors.

One cannot escape this effect by simply reporting lots of boxes, because doing so will require the classifier to be very good at rejecting false positives.  Instead, one wants the box proposal process to not miss high IOU boxes, without wild speculation. We offer a relatively straightforward strategy.  We train multiple RPN's to be somewhat redundant (so that if one RPN misses a high IOU box, another will get it), without overpredicting.   In what follows, we demonstrate how to do so and show how to balance redundancy against overprediction.  

Our contributions are three-fold: (1) We identify an important effect in few-shot object detection that causes existing detectors to have weaker than necessary performance in the few-shot regime. (2) We propose to overcome the proposal neglect effect by utilizing RPN redundancy. (3) We design an RPN ensemble mechanism that trains multiple RPN's simultaneously while enforcing diversity and cooperation among RPN's. We achieve state-of-the-art performance on COCO and PASCAL VOC in very few-shot settings. 

\section{Background}
\label{sec:related}

\textbf{Object Detection with Abundant Data.} The best-performing modern detectors are based on convolutional neural networks. There are two families of architecture, both relying on the remarkable fact that one can quite reliably tell whether an image region contains an object independent of category \citep{EndresHoiem, Regionproposal}.   In serial detection, a proposal process (RPN/s in what follows) offers the classifier a selection of locations likely to contain objects, and the classifier labels them, with the advantage that the classifier ``knows'' the likely support of the object fairly accurately.   This family includes R-CNN and its variants 
(for R-CNN \citep{girshick2014rich}; Fast R-CNN \citep{girshick2015fast}; Faster R-CNN \citep{ren2015faster}; Mask R-CNN \citep{he2017mask}; SPP-Net \citep{he2015spatial}; FPN \citep{lin2017feature}; and DCN \citep{dai2017deformable}).  In parallel, the proposal process and classification process are independent; these methods can be faster, but the classifier ``knows'' very little about the likely support of the object, which can affect accuracy.  This family includes YOLO and its variants (for YOLO versions \citep{redmon2016you, redmon2017yolo9000, redmon2018yolov3, bochkovskiy2020yolov4}; SSD \citep{liu2016ssd}; Cornernet \citep{law2018cornernet}; and ExtremeNet \citep{zhou2019bottom}).   This paper identifies an issue with the proposal process that can impede strong performance when there is very little training data (the {\em few-shot} case).  The effect is described in the context of serial detection, but likely occurs in  parallel detection too. 

\textbf{Few-Shot Object Detection.} Few-shot object detection involves detecting objects for which there are very few training examples. There is a rich few-shot classification literature (roots in~\citep{Thrun1998,fei2006one}). \citet{nikita2019div} uses ensemble procedures for few-shot classification. As to detection, \citet{chen2018lstd} proposes a regularized fine-tuning approach to transfer knowledge from a pre-trained detector to a few-shot detector. \citet{schwartz2018repmet} exploits metric-learning for modeling multi-modal distributions for each class. State-of-the-art few-shot detectors are usually serial~\citep{WangRH19, yan2019metarcnn, wang2020few, fan2020fsod, wu2020mpsr, Xiao2020FSDetView}.  The existing literature can be seen as variations on a standard framework, where one splits data into two sets of categories: base classes $C_b$ (which have many training examples) and novel classes $C_n$ (which have few).  The RPN and classifier are then trained with instances from the base classes, producing a detector for $|C_b|$ categories.  The final layer of the resulting classifier is expanded to classify into $|C_b|+|C_n|$ classes by inserting random weights connecting the final feature layer to the $|C_n|$ novel categories.  Now the model is fine-tuned using either only the novel class instances or a balanced dataset containing training instances of both base and novel classes. \citet{wang2020few} shows that a simple two-stage fine-tuning approach outperforms other complex methods. Much work seeks improvements by applying few-shot classification techniques. \citet{kang2018few} designs a meta-model that learns to reweight pre-trained features given few-shot data. \citet{WangRH19} and \citet{yan2019metarcnn} further explore the meta-learning direction by attaching meta-learned classifiers to Faster R-CNN. \citet{wu2020mpsr} improves few-shot detection by positive sample refinement.

Relatively little work adjusts the proposal process, which is usually seen as robust to few-shot issues because there are many base examples.  
One possibility is to introduce attention mechanisms and feed category-aware features instead of plain image features into the proposal process \citep{Hsieh19AttenFew, fan2020fsod, Xiao2020FSDetView, osokin20os2d}, as well as re-ranking proposals based on similarity with query images \citep{Hsieh19AttenFew, fan2020fsod}. Making the RPN category-aware improves the quality of novel class proposals, {\em but at inference time the model suffers from catastrophic forgetting of base categories} -- current category-aware features cannot summarize the very large number of base class examples efficiently or accurately.  An RPN that is generally well-behaved can still create serious trouble in the few-shot case by missing high IOU proposals for the novel classes during fine-tuning -- the {\em proposal neglect} effect. We show that this problem is severe in the few-shot regime, and can be fixed by a carefully constructed ensemble of RPNs without substantial loss of performance for the base classes.

\textbf{Evaluating Few-Shot Detectors.} The standard procedure is to compute average precision (AP) separately for novel and base categories for a detector that is engineered to detect $|C_b|+|C_n|$ classes, typically using standard test/train splits and standard novel/base splits \citep{wang2020few}. 
This evaluation procedure is the same as in incremental few-shot detection \citep{Yang2020ContextTransformerTO, rua2020incremental}. 
This procedure makes sense, because in most applications an incoming test image could contain instances from both base and novel classes.   Furthermore, the standard procedure exposes any catastrophic forgetting that occurs.    However, other evaluation methodologies occur, and some detectors are evaluated using variant procedures, making the comparison of AP's difficult. In one variant, the detector detects only the $|C_n|$ novel classes or only one novel class. In this paper, all reported results are for the standard procedure; when relevant, we re-implement and re-evaluate using the standard procedure. 



\section{Our Approach}
\label{method}
We believe that the proposal neglect effect is generic, and it applies to any detector that uses a structure like the standard structure. For this reason, we focus on finding and fixing the effect within a standard state-of-the-art few-shot object detection framework, as described below. 

\begin{figure}[t!]
\centering\includegraphics[width=.8\linewidth]{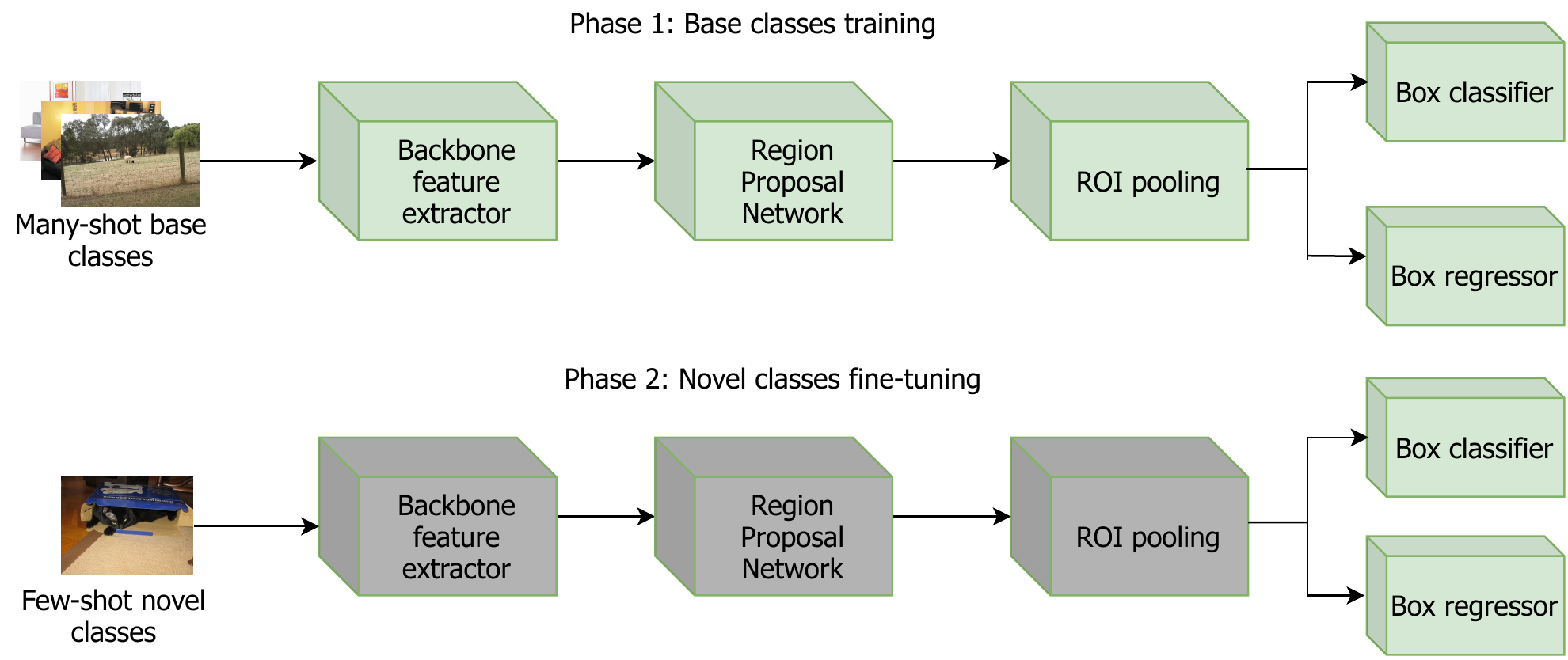}
\vspace{1mm}
\caption{Illustration of two-phase training in Faster R-CNN. During phase 1, all blocks are trained. During phase 2, only the top-layer classifier and bounding box regressor are fine-tuned.}
\label{two_stage}
\end{figure}

\textbf{Few-Shot Object Detection Framework.}
\label{sec:method}
We use the few-shot detection setting introduced in \citet{kang2018few}. We split the dataset into two sets of categories: base classes $C_b$ and novel classes $C_n$. As shown in Figure \ref{two_stage}, the training process is two-phase: (1) base classes training, and (2) fine-tuning with novel classes. In phase 1, the model is trained with base class instances which results in a $|C_b|$-way detector. After base classes training, weights for novel classes are randomly initialized, making the classifier a $(|C_b| + |C_n|)$-way classifier. In phase 2, the model is fine-tuned using either a set of few novel class instances or a balanced dataset containing both novel and base classes. After the fine-tuning phase, we evaluate our model by average precision (AP) on novel and base categories. Although the focus of few-shot detection is the novel classes, since most test images contain instances from both base and novel classes, it is essential to maintain good performance on base classes. 

\begin{figure}[t!]
\centering
\includegraphics[width=.9\linewidth]{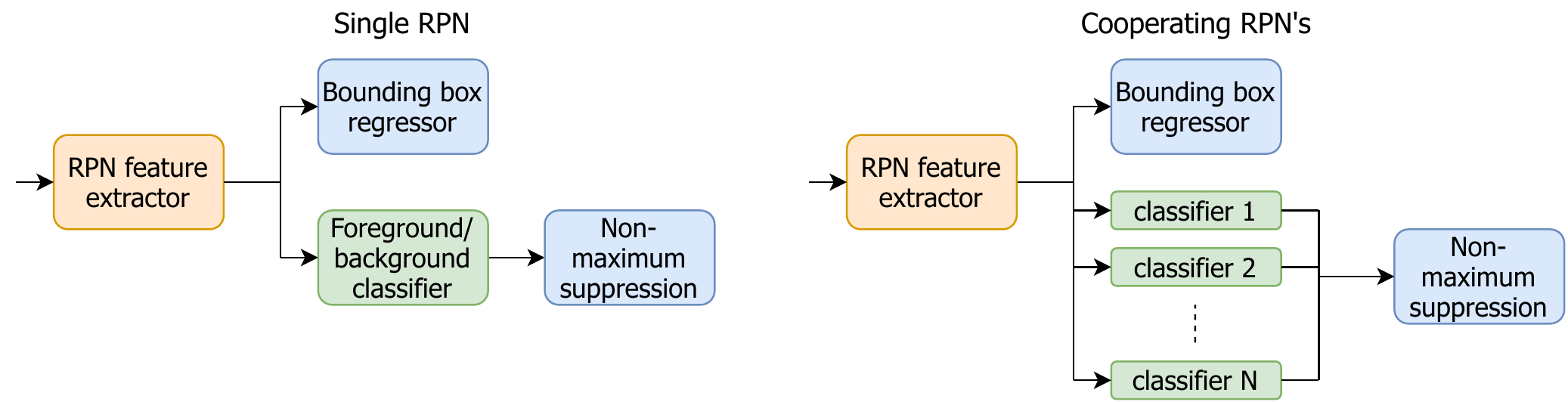}
\vspace{1mm}
\caption{{\bf Left} is the original structure for the RPN in Faster R-CNN;  {\bf right} is our CoRPNs, consisting of cooperating box classifiers.}
\label{rpn}
\end{figure}

We adopt the widely-used Faster R-CNN~\citep{ren2015faster} as our base model. As shown in Figure~\ref{two_stage}, Faster R-CNN is a serial detector, which consists of a backbone image feature extractor, a regional proposal network (RPN), followed by the region of interest (ROI) pooling layer, and a bounding box classifier and a bounding box regressor on top of the model. The RPN determines if a box is a foreground or a background box. Following the RPN is non-maximum suppression (NMS) which ranks and selects top proposal boxes. After passing the ROI pooling layer, the predictor's head classifies and localizes each box. In phase 1, the whole model is trained on many-shot base class instances. Phase 2 fine-tunes the head classifier and the bounding box regressor with novel class instances; everything before and including the proposal generator is frozen.

\textbf{Learning Cooperating RPN's (CoRPNs).}
We wish to avoid high IOU boxes for novel classes being dropped by RPN when novel classes are trained. Our strategy is to train multiple redundant RPN's.  The RPN's should be distinct, but cooperate -- if one misses a high IOU box, we want another to find it.  However, they should not be so distinct that the classifier is flooded with false positives.

As Figure \ref{rpn} shows, Faster R-CNN's RPN consists of a feature extractor, a binary classifier (which decides whether a box is foreground or background), and a bounding box regressor (which is not relevant to our current purpose). There is no reason for our RPN's to use distinct sets of features, and we do not want to create problems with variance, so we construct redundant classifiers while keeping both the feature extractor and the bounding box regressor shared between all RPN's. In what follows, a reference to an RPN is actually a reference to the RPN classifier, unless otherwise noted. An RPN with a single classifier is trained with a cross-entropy loss $\mathcal{L}_{cls} = \mathcal{L}_{CE}$ and produces a single prediction.  In our case, we train $N$ different binary classifiers simultaneously, and must determine (1) what prediction is made at test time and (2) what gradient goes to what classifier at training time. At test time, a given box gets the score from the most certain RPN. If the highest foreground probability is closer to one than the highest background probability, the box is foreground; otherwise, it is background. 

Training time is more interesting. Merely taking the gradient of the best RPN score is not good enough, because we may find that one RPN scores all boxes, and the others do nothing interesting. For any foreground box, we want at least one RPN to have a very strong foreground score {\em and} all others to have good foreground scores too (so that no foreground box is missed).  

We use the following strategy. For a specific anchor box $i$, each RPN $j$ outputs a raw score $r_i^j$, indicating if the box is a foreground box or not: $r_i = [r^1_i, r^2_i,\dots, r^N_i].$ After applying a sigmoid, the $j$th RPN produces the foreground probability $f^j_i = \sigma(r^j_i)$ for anchor box $i$. We choose the score from the $j^*$th RPN such that $j^* = {\mathrm{argmin}_j} \min\{f^j_i, 1 - f^j_i\}$, namely the most certain RPN which produces probability closest to the edge of the $[0, 1]$ interval. At training time, only the chosen $j^*$th RPN gets the gradient from anchor box $i$. The RPN selection procedure is per-box, even adjacent boxes can pass through different RPN's. 

Other than the standard cross-entropy loss, we use two additional loss terms: a diversity loss $L_{div}$ encourages RPN's to be distinct, and a cooperation loss $L_{coop}$ encourages cooperation and suppresses foreground false negatives. The final loss $\mathcal{L}_{cls} := \mathcal{L}^{j^*}_{CE} + \lambda_{d} \mathcal{L}_{div} + \lambda_{c} \mathcal{L}_{coop}$, where $\lambda_{d}$ and $\lambda_{c}$ are trade-off hyperparameters.

\textbf{Enforcing Diversity.}
We do not want our RPN's to be too similar. For each positive anchor box, RPN responses should be different because we want ensure that at novel class training time, if one RPN misses a high IOU anchor box, another will find it. To this end, we propose a methodology to enforce diversity among RPN's. Given a set of $N_A$ anchor boxes, the $N$ RPN's produce an $N$ by $N_A$ matrix of probabilities $\mathcal{F} = [f^1, f^2, \dots, f^N]^T$. The covariance matrix $\Sigma(\mathcal{F})$ is $\Sigma(\mathcal{F}) = \operatorname{E} [(f^j-\operatorname{E} [f^j])(f^k-\operatorname{E} [f^k])]$. 
We define the diversity loss $\mathcal{L}_{div}$ by the log determinant loss $\mathcal{L}_{div} := -\log(\det(\Sigma(\mathcal{F})))$. 

By the diversity loss, we encourage the probability matrix to have rank $N$, so each RPN is reacting differently on the collection of $N_A$ boxes. This procedure ensures each RPN to be the most certain RPN for some boxes, so that every RPN is being selected and trained. Omitting this loss can cause an RPN to receive no or little training.

\textbf{Learning to Cooperate.}
We also want the RPN's to cooperate so that they all agree to a certain extent for foreground boxes. We propose a cooperation loss to prevent any RPN from firmly rejecting any foreground box. For foreground box $i$, with the $j$th RPN, we define the cooperation loss $\mathcal{L}_{coop}^{i, j} :=\max\{0, \phi - f^{j}_i\}$, where $\phi$ is a constant parameter (usually less than 0.5), acting as a lower bound for each RPN's probability assigning to a foreground box. If a RPN's response is below $\phi$, that RPN is going to be penalized. The final cooperation loss is an average of cooperation losses over all foreground boxes and all RPN's.

\section{Experiments}
\label{sec:expt}

\textbf{Datasets and Implementation Details.} We evaluate our approach on two widely-used few-shot detection benchmarks: MS-COCO \citep{lin2014microsoft} and PASCAL VOC (07 + 12) \citep{everingham2010pascal}. For a fair comparison, we use the same train/test splits and novel class instances as in \citet{kang2018few, wang2020few} to learn and evaluate all models. On COCO, we report base/novel classes AP, AP50, and AP75 under shots 1, 2, 3, 5, 10, and 30. On PASCAL VOC, we report AP50 for three different base/novel classes splits under shots 1, 2, 3, 5, and 10. Consistent with recent works \citep{yan2019metarcnn, wang2020few}, we mainly use an ImageNet pretrained \citep{russakovsky2015imagenet} ResNet-101 architecture \citep{he2016deep} as the backbone for all models, unless otherwise noted. More implementation details are included in the Appendix.



\textbf{Baselines and Evaluation Procedure.} To investigate the {\em proposal neglect} effect and for a fair comparison, we mainly focus on comparing against the state-of-the-art baseline TFA \citep{wang2020few}. Our approach introduces CoRPNs into TFA, while keeping other model components and design choices unchanged. In addition, we thoroughly compare with a variety of recent few-shot detectors, including CoAE~\citep{Hsieh19AttenFew}, Meta R-CNN~\citep{yan2019metarcnn}, FSOD~\citep{fan2020fsod}, MPSR~\citep{wu2020mpsr}, FSDetView~\citep{Xiao2020FSDetView}, and ONCE~\citep{rua2020incremental}. These baselines address other aspects of few-shot detection which are different from us (Section~\ref{sec:related}), and their modifications are thus largely orthogonal to our effort here. Note that our evaluation follows the standard procedure in~\citet{wang2020few}. As explained in Section~\ref{sec:related}, this standard procedure computes AP separately for novel and base categories for a detector that is engineered to detect $|C_b|+|C_n|$ classes. For a fair comparison, we re-evaluate CoAE~\citep{Hsieh19AttenFew} and FSOD~\citep{fan2020fsod} using the standard procedure.

\textbf{Main Results.} Tables~\ref{tab:coco_novel} and~\ref{tab:voc_novel} summarize the results for novel classes on COCO and PASCAL VOC, respectively, and Table~\ref{tab:coco_base} summarizes the results for base classes on COCO.

\textbf{\textit{Comparisons with the Main Baseline TFA on Novel Classes.}} From Tables~\ref{tab:coco_novel} and~\ref{tab:voc_novel}, we have the following important observations. (\RNum{1}) Our approach produces a substantial improvement in AP over TFA~\citep{wang2020few} on novel classes {\em in the very low-shot regime} (1, 2, and 3 shots), and marginal improvement or sometimes slight degradation in the higher-shot regime. These improvements are manifest for both existing benchmarks. Interestingly, on the more challenging COCO dataset, our improvements over TFA are {\em consistent across different shots} (except slight degradation in 30 shot under AP75 with fully-connected classifier). (\RNum{2}) We investigate two types of classifiers: either fully-connected (denoted as `fc' in Tables~\ref{tab:coco_novel} and~\ref{tab:voc_novel}) or cosine (denoted as `cos' in Tables~\ref{tab:coco_novel} and~\ref{tab:voc_novel}). Note that our approach obtains improvements regardless of classifier choice. This is because CoRPNs is a strategy to control variance in the estimate of classifier parameters {\em that applies independently of the classifier}. Any high IOU box missing from the RPN output in the training phase must cause variance for the few-shot regime. Because there are very few such boxes, the effect of not losing boxes is pronounced. We provide visualization comparisons of detection results in the Appendix.

\textbf{\textit{Comparisons with Other State-of-the-Art Approaches.}} With our simple modification on RPN, we also outperform other sophisticated approaches on both benchmarks in the very low-shot regime, and achieve comparable performance in the higher-shot regime. In particular, we significantly outperform those baselines that introduce attention mechanisms for adjusting proposal generation~\citep{Hsieh19AttenFew,fan2020fsod}. For other approaches that improve few-shot detection from different perspectives, such as exploiting better multi-scale representation \citep{wu2020mpsr}, our approach can be potentially combined with them for further improvements.

\textbf{\textit{Comparisons on Base Classes.}} While improving detection on novel classes through fine-tuning, we maintain strong performance on base classes {\em without suffering from catastrophic forgetting} as shown in Table~\ref{tab:coco_base}. By contrast, the performance of the state-of-the-art baselines dramatically drops, demonstrating that they cannot simultaneously deal with both novel and base classes.
\def\Hyphen{{\hspace{1.8mm}} - {\hspace{1.8mm}}}
\begin{table}[hbt!]
\centering
\renewcommand{\arraystretch}{1.1} 
\resizebox{\textwidth}{!}{
\begin{tabular}{lllccc|ccc|ccc}


&&&\multicolumn{3}{c}{1-shot} & \multicolumn{3}{c}{2-shot} & \multicolumn{3}{c}{3-shot} \\
& Method &\; Backbone &\; AP & AP50 & AP75 & AP & AP50 & AP75 & AP & AP50 & AP75 \\
\hline
\multirow{3}{*}{Ours} & CoRPNs w/ fc &\;ResNet-101 &\;{3.4} & {6.7} & {3.0} & \textcolor{red}{\bf5.4} & \textcolor{red}{\bf10.4} & \textcolor{blue}{\emph{5.1}} & \textcolor{red}{\bf7.1} & \textcolor{red}{\bf13.7} & \textcolor{blue}{\emph{6.8}} \\
& CoRPNs w/ cos &\;ResNet-101 &\;\textcolor{red}{\bf4.1} & \textcolor{blue}{\emph{7.2}} & \textcolor{red}{\bf4.4} & \textcolor{red}{\bf5.4} & {9.6} & \textcolor{red}{\bf5.6} & \textcolor{red}{\bf7.1} & \textcolor{blue}{\emph{13.2}} & \textcolor{red}{\bf7.2} \\
& CoRPNs w/ cos &\;ResNet-50 &\;\textcolor{blue}{\emph{3.7}} & {6.8} & \textcolor{blue}{\emph{3.8}} & {4.7} & {8.8} & {4.4} & {6.3} & {12.0} & {6.0}\\
\hline
\multirow{2}{*}{Main baselines} & TFA w/ fc \citep{wang2020few} &\;ResNet-101 &\;{2.9} & {5.7} & {2.8} & {4.3}& {8.5} & {4.1} & {6.7} & {12.6} & {6.6} \\
& TFA w/ cos \citep{wang2020few} &\;ResNet-101 &\;{3.4} & {5.8} & \textcolor{blue}{\emph{3.8}} & {4.6}& {8.3} & {4.8} & {6.6} & {12.1} & {6.5} \\
\hline
\multirow{6}{*}{Other baselines} & Meta R-CNN \citep{yan2019metarcnn} &\;ResNet-101 &\;{--} & {--} & {--} & {--} & {--} & {--} & {--} & {--} & {--}\\
& FRCN+ft-full \citep{wang2020few} &\;ResNet-101 &\;{--} & {--} & {--} & {--} & {--} & {--} & {--} & {--} & {--}\\
& MPSR \citep{wu2020mpsr}&\;ResNet-101 &\;{2.3} & {4.1} & {2.3} & {3.5} & {6.3} & {3.4} & {5.2} & {9.5} & {5.1}\\
& FsDetView \citep{Xiao2020FSDetView}&\;ResNet-101 &\;{2.9} & \textcolor{red}{
\bf8.3} & {1.2} & {3.7} & \textcolor{blue}{\emph{10.3}} & {1.6} & {4.7} & {12.9} & {2.0}\\
& ONCE \citep{rua2020incremental}&\;ResNet-50 &\;{0.7} & {--} & {--} & {--} & {--} & {--} & {--} & {--} & {--}\\
& FSOD* \citep{fan2020fsod}&\;ResNet-50 &\;{2.4} & {4.8} & {2.0} & {2.9} & {5.9} & {2.7} & {3.7} & {7.2} & {3.3}\\
\hline
&&&\multicolumn{3}{c}{5-shot} & \multicolumn{3}{c}{10-shot} & \multicolumn{3}{c}{30-shot} \\ 
& Method &\; Backbone &\; AP & AP50 & AP75 & AP & AP50 & AP75 & AP & AP50 & AP75 \\
\hline
\multirow{3}{*}{Ours} & CoRPNs w/ fc &\;ResNet-101 &\;\textcolor{red}{\bf8.9} & \textcolor{red}{\bf16.9} & \textcolor{blue}{\emph{8.6}}& \textcolor{blue}{\emph{10.5}} & \textcolor{red}{\bf20.2} & \textcolor{blue}{\emph{9.8}}& {13.5} & \textcolor{blue}{\emph{25.0}} & {12.9}\\
& CoRPNs w/ cos &\;ResNet-101 &\;\textcolor{blue}{\emph{8.8}} & \textcolor{blue}{\emph{16.4}} & \textcolor{red}{\bf8.7}& \textcolor{red}{\bf10.6} & \textcolor{blue}{\emph{19.9}} & \textcolor{red}{\bf10.1}& \textcolor{red}{\bf13.9} & \textcolor{red}{\bf25.1} & \textcolor{red}{\bf13.9}\\
& CoRPNs w/ cos &\;ResNet-50 &\;{7.8} & {14.4} & {7.6} & {9.0} & {17.6} & {8.3} & {13.4} & {24.6} & {13.3}\\
\hline
\multirow{2}{*}{Main baselines} & TFA w/ fc \citep{wang2020few} &\;ResNet-101 &\;{8.4}& {16.0} & {8.4} & {10.0}& {19.2} & {9.2} & {13.4}& {24.7} & {13.2} \\
& TFA w/ cos \citep{wang2020few} &\;ResNet-101 &\;{8.3}& {15.3} & {8.0} & {10.0}& {19.1} & {9.3} & \textcolor{blue}{\emph{13.7}}& {24.9} & \textcolor{blue}{\emph{13.4}} \\
\hline
\multirow{5}{*}{Other baselines} & Meta R-CNN \citep{yan2019metarcnn} &\;ResNet-101 &\;{--} & {--} & {--} & {8.7} & {--} & {6.6} & {12.4} & {--} & {10.8}\\
& FRCN+ft-full \citep{wang2020few} &\;ResNet-101 &\;{--} & {--} & {--} & {9.2} & {--} & {9.2} & {12.5} & {--} & {12.0}\\
& MPSR \citep{wu2020mpsr}&\;ResNet-101 &\;{6.7} & {12.6} & {6.4} & {9.7} & {18.0} & {9.4} & \textcolor{blue}{\emph{13.7}} & \textcolor{blue}{\emph{25.0}} & \textcolor{blue}{\emph{13.4}}\\
& FsDetView \citep{Xiao2020FSDetView}&\;ResNet-101 &\;{5.8} & {15.6} & {2.9} & {6.7} & {17.3} & {3.7} & {9.6} & {22.1} & {6.6} \\ 
& ONCE \citep{rua2020incremental}&\;ResNet-50 &\;{1.0} & {--} & {--} & {1.2} & {--} & {--} & {--} & {--} & {--}\\
& FSOD* \citep{fan2020fsod}&\;ResNet-50 &\;{4.2} & {8.2} & {4.0} & {4.3} & {8.7} & {3.8} & {5.4} & {10.4} & {5.0}\\

\end{tabular}
}
\vspace{0.1cm}
\caption{Few-shot detection performance on COCO novel classes. The upper row shows the 1, 2, 3-shot results, and the lower row shows the 5, 10, 30-shot results. Results in \textcolor{red}{\bf{red}} are the best, and results in \textcolor{blue}{\emph{blue}} are the second best. All approaches are evaluated following the standard procedure in \citet{wang2020few}. *Model re-evaluated using the standard procedure for a fair comparison. `--' denotes that numbers are not reported in the corresponding paper.  Note that the publicly released models of ONCE and FSOD are based on ResNet-50; we include our CoRPNs based on ResNet-50 as well for a fair comparison. CoRPNs consistently outperform state of the art in almost all settings, {\em with substantial improvements especially in the very few-shot regime}. Our strategy is also effective {\em regardless of classifier choice}.}
\vspace{-4mm}
\label{tab:coco_novel}
\end{table}

\begin{table}[hbt!]
\centering
\renewcommand{\arraystretch}{1.2} 
\resizebox{\textwidth}{!}{
\begin{tabular}{llccccc|ccccc|ccccc}
&& \multicolumn{5}{c}{Novel Set 1} & \multicolumn{5}{c}{Novel Set 2} & \multicolumn{5}{c}{Novel Set 3}\\
& Method & shot=1 & 2 & 3 & 5 & 10& shot=1 & 2 & 3 & 5 & 10 & shot=1 & 2 & 3 & 5 & 10\\
\hline
\multirow{2}{*}{Ours} & CoRPNs w/ fc (Ours) &\;{40.8} &\textcolor{red}{\bf44.8} & {45.7} & {53.1} & {54.8} & {20.4} &{29.2} & {36.3} & {36.5} & {41.5} & {29.4} & {40.4} & \textcolor{red}{\bf44.7} & \textcolor{red}{\bf51.7} & \textcolor{blue}{\emph{49.9}}\\
& CoRPNs w/ cos (Ours) &\;\textcolor{red}{\bf44.4} & {38.5} & \textcolor{blue}{\emph{46.4}} & {54.1} & {55.7} & \textcolor{red}{\bf25.7} &\textcolor{red}{\bf29.5} & \textcolor{blue}{\emph{37.3}} & {36.2} & {41.3} & \textcolor{red}{\bf35.8} & \textcolor{red}{\bf41.8} & \textcolor{blue}{\emph{44.6}} & \textcolor{blue}{\emph{51.6}} & {49.6}\\
\hline
\multirow{2}{*}{Main baselines} & TFA w/ fc (baseline) \citep{wang2020few} &\;{36.8} & {29.1} & {43.6} & \textcolor{red}{\bf55.7} & {57.0} & {18.2} & {29.0} & {33.4} & {35.5} & {39.0} & {27.7} & {33.6} & {42.5} & {48.7} & \textcolor{red}{\bf50.2}\\
& TFA w/ cos (baseline) \citep{wang2020few} &\;{39.8} & {36.1} & {44.7} & \textcolor{red}{\bf55.7} & {56.0} & {23.5} & {26.9} & {34.1} & {35.1} & {39.1} & {30.8} & {34.8} & {42.8} & {49.5} & {49.8}\\
\hline
\multirow{5}{*}{Other baselines} & FRCN+ft-full \citep{wang2020few} &\;{15.2} & {20.3} & {29.0} & {40.1} & {45.5} & {13.4} & {20.6} & {28.6} & {32.4} & {38.8} & {19.6} & {20.8} & {28.7} & {42.2} & {42.1}\\
& Meta R-CNN \citep{yan2019metarcnn} &\;{19.9} & {25.5} & {35.0} & {45.7} & {51.5} & {10.4} & {19.4} & {29.6} & {34.8} & {45.4} & {14.3} & {18.2} & {27.5} & {41.2} & {48.1}\\
& CoAE* \citep{Hsieh19AttenFew}&\;{12.7} & {14.6} & {14.8} & {18.2} & {21.7} & {4.4} & {11.3} & {20.5} & {18.0} & {19.0} & {6.3} & {7.6} & {9.5} & {15.0} & {19.0}\\
& MPSR \citep{wu2020mpsr}&\;\textcolor{blue}{\emph{41.7}} & \textcolor{blue}{\emph{43.1}} & \textcolor{red}{\bf51.4} & {55.2} & \textcolor{red}{\bf61.8} & \textcolor{blue}{\emph{24.4}} & \textcolor{red}{\bf29.5} & \textcolor{red}{\bf39.2} & \textcolor{red}{\bf39.9} & \textcolor{red}{\bf47.8} & \textcolor{blue}{\emph{35.6}} & \textcolor{blue}{\emph{40.6}} & {42.3} & {48.0} & {49.7}\\
& FsDetView \citep{Xiao2020FSDetView}&\;{24.2} & {35.3} & {42.2} & {49.1} & \textcolor{blue}{\emph{57.4}} & {21.6} & {24.6} & {31.9} & \textcolor{blue}{\emph{37.0}} & \textcolor{blue}{\emph{45.7}} & {21.2} & {30.0} & {37.2} & {43.8} & {49.6}\\
\end{tabular}
}
\vspace{0.1cm}
\caption{Few-shot detection performance (AP50) on PASCAL VOC novel classes under three base/novel splits. All models are based on Faster R-CNN with a ResNet-101 backbone. Results in \textcolor{red}{\bf{red}} are the best, and results in \textcolor{blue}{\emph{blue}} are the second best. We follow the standard evaluation procedure in \citet{wang2020few}. *Model re-evaluated under the standard procedure. CoRPNs outperform all the baselines in the very low shots, and achieve comparable performance in the higher shots.}
\vspace{-4mm}
\label{tab:voc_novel}
\end{table}

\begin{table}[hbt!]
\centering
\renewcommand{\arraystretch}{1.2} 
\resizebox{\textwidth}{!}{
\begin{tabular}{lllccc|ccc|ccc}
&&&\multicolumn{3}{c}{1-shot finetuned} & \multicolumn{3}{c}{2-shot finetuned} & \multicolumn{3}{c}{3-shot finetuned} \\
& Method &\; Backbone &\; AP & AP50 & AP75 & AP & AP50 & AP75 & AP & AP50 & AP75 \\
\hline
\multirow{2}{*}{Ours} & CoRPNs w/ cos &\;ResNet-101 &\;{34.1} & {55.1} & {36.5} & {34.7} & {55.3} & {37.3} & {34.8} & {55.2} & {37.6}\\
& CoRPNs w/ cos &\;ResNet-50 &\;{32.1} & {52.9} & {34.4} & {32.7} & {52.9} & {35.5} & {32.6} & {52.4} & {35.4}\\
\hline
\multirow{1}{*}{Main baseline} & TFA w/ cos \citep{wang2020few} &\;ResNet-101 &\;{34.1} & {54.7} & {36.4} & {34.7} & {55.1} & {37.6} & {34.7} & {54.8} & {37.9}\\
\hline
\multirow{3}{*}{Other baselines} & MPSR \citep{wu2020mpsr}&\;ResNet-101 &\;{12.1} & {17.1} & {14.2} & {14.4} & {20.7} & {16.9} & {15.8} & {23.3} & {18.3}\\
& FsDetView \citep{Xiao2020FSDetView}&\;ResNet-101 &\;{1.9} & {5.7} & {0.8} & {2.7} & {8.2} & {0.9} & {3.9} & {10.8} & {2.0}\\
& FSOD* \citep{fan2020fsod}&\;ResNet-50 &\;{11.9} & {20.3} & {12.5} & {15.6} & {24.4} & {17.2} & {17.4} & {27.3} & {19.0}\\
\hline

&&&\multicolumn{3}{c}{5-shot finetuned} & \multicolumn{3}{c}{10-shot finetuned} & \multicolumn{3}{c}{30-shot finetuned} \\ 
& Method &\; Backbone &\; AP & AP50 & AP75 & AP & AP50 & AP75 & AP & AP50 & AP75 \\
\hline
\multirow{2}{*}{Ours} & CoRPNs w/ cos &\;ResNet-101 &\;{34.7} & {54.8} & {37.5} & {34.6} & {54.5} & {38.2} & {35.8} & {55.4} & {39.4}\\
& CoRPNs w/ cos &\;ResNet-50 &\;{32.3} & {51.6} & {34.9} & {32.7} & {51.9} & {36.0} & {33.5} & {52.7} & {37.0}\\
\hline
\multirow{1}{*}{Main baseline} & TFA w/ cos \citep{wang2020few} &\;ResNet-101 &\;{34.7} & {54.4} & {37.6} & {35.0} & {55.0} & {38.3} & {35.8} & {55.5} & {39.4}\\
\hline
\multirow{3}{*}{Other baselines} & MPSR \citep{wu2020mpsr}&\;ResNet-101 &\;{17.4} & {25.9} & {19.8} & {19.5} & {29.5} & {21.9} & {21.0} & {32.4} & {23.4}\\
& FsDetView \citep{Xiao2020FSDetView}&\;ResNet-101 &\;{5.3} & {14.2} & {2.8} & {6.4} & {15.9} & {4.1} & {9.0} & {20.6} & {6.7}\\
& FSOD* \citep{fan2020fsod}&\;ResNet-50 &\;{16.7} & {26.2} & {18.3} & {18.9} & {29.3} & {20.7} & {18.8} & {29.4} & {20.1}\\
\end{tabular}
}
\vspace{0.1cm}
\caption{Detection performance on COCO base classes after {\em fine-tuning with k-shot novel classes instances.} *Model re-evaluated under the standard procedure~\citep{wang2020few}. Our CoRPNs and TFA~\citep{wang2020few} maintain good performance on base classes, whereas MPSR~\citep{wu2020mpsr} and FsDetView~\citep{Xiao2020FSDetView} suffer from severe catastrophic forgetting.}
\vspace{-4mm}
\label{tab:coco_base}
\end{table}
\textbf{Ablation Studies.} We conduct a series of ablations to evaluate the contribution of each component and different design choices. Specifically, we show that (1) with our novel loss terms, CoRPNs outperform naive RPN ensembles; (2) CoRPNs outperform an existing cosine loss based diversity method; (3) our cooperation loss is helpful; (4) the number of RPN's matters. 

\textbf{\textit{CoRPNs vs. Naive RPN Ensembles.}}
We compare CoRPNs with an ensemble of the same number of RPN's trained separately, each with a different initialization. At the fine-tuning phase, we apply the same RPN selection mechanism to both approaches. Table~\ref{tab:naive} shows that novel AP50 of CoRPNs largely outperforms naive ensembles. This suggests that {\em pure redundancy does not lead to improvements}, confirming that our loss terms enforce diversity and cooperation. In addition, there are $N$ groups of parameters in naive ensembles, because each model is trained separately. Hence, the number of parameters and training time of our CoRPNs are $N$ times less than naive ensembles.

\textbf{\textit{CoRPNs vs. Cosine Loss Diversity.}}
Results in Table~\ref{tab:nikita} confirm that in this context, our diversity loss -- the log-determinant loss -- outperforms another diversity loss introduced in \citet{nikita2019div}.  \citet{nikita2019div} encourage diversity among classifiers by enlarging the cosine distance in classifier outputs. To compare with their method, we replace our diversity loss term by a pairwise cosine similarity loss among RPN's (if there are $N$ RPN's, the pairwise cosine similarity loss is an average over all {\small${N(N-1)}/{2}$} pairs).

\textbf{\textit{Avoiding False Negatives.}}
In our cooperation loss $\mathcal{L}_{coop}$ (which pushes all RPN's to agree on certain degree for foreground boxes) has a threshold hyperparameter $\phi$. Table \ref{table:thres} reports results under different thresholds on PASCAL VOC novel split 3 fine-tuned on 1-shot novel instance. In addition to AP50, we report the average number of false negative foreground boxes (`Avg \# FN') during fine-tuning and the average number of foreground samples (`Avg \# FG')  after non-maximum suppression (NMS). The table shows that the cooperation loss term causes the number of false negatives (i.e. high IOU boxes not presented to the classifier at fine-tuning time) to decrease. Furthermore, the average number of foreground samples after NMS increases. 

\begin{minipage}{\textwidth}
\centering
\begin{minipage}[b]{0.45\textwidth}
\centering
\renewcommand{\arraystretch}{1.2}
\renewcommand{\tabcolsep}{1.2mm}
\resizebox{0.82\linewidth}{!}{\begin{tabular}{ll|ll}
    Method & AP50 & Method & AP50 \\
    \hline
    2 RPN's Naive &  24.6    & 5 RPN's Naive &  23.7 \\
    2 RPN's Ours & \textbf{35.8} & 5 RPN's Ours & \textbf{34.8}  \\
  \end{tabular}}%
 \captionof{table}{CoRPNs significantly outperform naive ensembles of RPN's.  The table shows novel class AP50 after phase 2 of our (resp. naive) RPN ensembles with 2 or 5 RPN's, all trained and evaluated under PASCAL VOC novel split 3, shot 1, using the same parameter settings. }%
 \label{tab:naive}
 \renewcommand{\arraystretch}{1.2}
\renewcommand{\tabcolsep}{1.2mm}
\resizebox{\linewidth}{!}{\begin{tabular}{lll}
    & Method & AP50 \\
    \midrule
     & 2 RPN's, Div and Coop \citep{nikita2019div} &  32.4    \\
    & 2 RPN's, CoRPNs (Ours) & \textbf{35.8}   \\
  \end{tabular}}
\captionof{table}{Our diversity enforcing term -- the log-determinant loss -- offers improvements over the pairwise cosine similarity based diversity loss in~\citet{nikita2019div}. The table shows novel class AP50 of both models, trained and evaluated under PASCAL VOC novel split 3, shot 1, using the same parameter settings.}
\label{tab:nikita}
 \end{minipage}
 \quad
\begin{minipage}[b]{0.49\textwidth}
\centering
  \resizebox{\columnwidth}{!}{%
  \begin{tabular}{lp{1.5cm}ccc}
    \multicolumn{2}{c}{Method} & AP50 & Avg \# FN ($\downarrow$)& Avg \# FG ($\uparrow$) \\
    \midrule
    \multicolumn{2}{c}{TFA \citep{wang2020few}} &  28.9  & 3.1 & 18.6\\
    \midrule
    \multirow{5}{*}{CoRPNs w/}&$\phi$ = 0.1 &  \bf{29.5}  & 2.4 & 22.3\\
    &$\phi$ = 0.3 &  \bf{31.5}  & 3.0 & 19.5\\ 
    &$\phi$ = 0.5 &  \bf{32.2}  & 2.5 & 19.3\\
  &$\phi$ = 0.7 &  26.8  & 1.3 & 21.3\\
    &$\phi$ = 0.9 & \bf{31.7}  & 0.8 & 20.0\\
  \end{tabular}
  }
\captionof{table}{Our threshold $\phi$ controls the average number of false-negative foreground boxes and the number of foreground samples. The table shows the novel class AP50, the average number of false-negative foreground boxes (with threshold = 0.5), and the average number of foreground boxes after NMS (higher is better). The last two numbers are calculated during the fine-tuning phase when RPN's are frozen. All models are trained and evaluated under PASCAL VOC novel split 3, shot 1. At phase 2, different from other experiments, we fine-tune with novel classes only, so the last two columns capture each model's ability to detect novel class boxes.}
\label{table:thres}
\end{minipage}
\end{minipage}

\begin{figure}[t]
\centering
\includegraphics[width=0.9\linewidth]{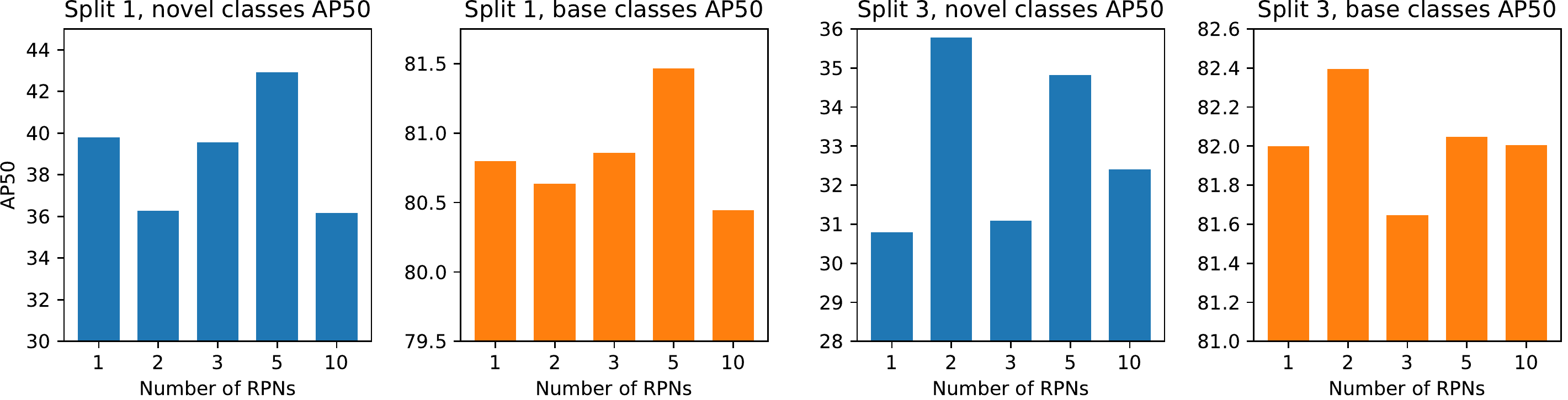}
\vspace{0.5mm}
\caption{Effect of the number of RPN's on novel and base AP50, under PASCAL VOC splits 1 and 3, both with shot 1 and fine-tuned on base + novel classes. Note that the horizontal scale is {\em not linear}.  Though there is considerable variance, the graphs suggest that at small numbers of RPN's, novel AP50 ({\bf blue}) is somewhat depressed, likely because some high IOU boxes are missed. As the number of RPN's increases, there is an improvement, but when there are a large number of RPN's, the rate at which false positive boxes passes the RPN system increases and the gain is lost.  The effect might be present for base classes as well ({\bf orange}), but is smaller (the vertical scale is different).}
\label{numRPNs}
\end{figure}

\textbf{\textit{The Number of RPN's.}} As expected, increasing the number of RPN's produces improvements up to a point, and then either the number of false positives or variance in the RPN parameter estimates leads to a decline in performance (Figure~\ref{numRPNs}). This figure plots the base AP50 (after phase 1) and novel AP50 under PASCAL VOC novel splits 1 and 3, shot 1. Note that the optimal number of RPN's differs over different splits (and assumably datasets). As shown in Figure \ref{numRPNs}, the optimal number is different between PASCAL VOC split 1 (left two figures) and split 3 (right two figures).
\section{Conclusion}
\label{sec:conclu}

A substantial improvement in few-shot performance can be obtained by 
engineering a system of RPN's to ensure that high IOU boxes for few-shot training images almost always pass the system of RPN's. Our method achieves a new state of the art on widely-used benchmarks and outperforms the current state of the art by a large margin in the very few-shot regime.   This is because, as ablation experiments indicate, proposal neglect is a real effect.  
We engineered our ensemble of RPN's by using a diversity loss and a cooperation loss, which gave us some control of 
the false negative rate.   We showed that controlling proposal neglect is an essential part of building a strong few-shot object detector by
showing that the state-of-the-art few-shot detector could be improved in this way.   We do not claim that our method is the best way to control proposal neglect.  For example, our $\phi$ threshold in the cooperation loss may cause false positives to be correlated.  We plan to investigate methods that can offer more delicate control of the false negative rate.  Furthermore, it is very likely that proposal  neglect applies to one-stage detectors like YOLO \citep{redmon2018yolov3}, and future work will investigate this.

\pagebreak



\bibliography{iclr2021_conference}
\bibliographystyle{iclr2021_conference}


\end{document}